\documentclass[10pt,twocolumn,letterpaper]{article}

\usepackage[pagenumbers]{cvpr} 

\usepackage{graphicx}
\usepackage{amsmath}
\usepackage{amssymb}
\usepackage{booktabs,subcaption,amsfonts,dcolumn}
\usepackage{dsfont}
\usepackage{times}
\usepackage{epsfig}
\usepackage{color, colortbl}
\usepackage{multirow}
\usepackage{enumitem}
\usepackage{amsmath}
\usepackage{makecell}
\usepackage{hhline}
\usepackage{cellspace}
\usepackage{bbding}
\usepackage{bm}
\usepackage{nicefrac}     
\usepackage{microtype}     
\usepackage{verbatim}
\usepackage{color}
\usepackage{float}
\usepackage{enumitem}
\usepackage{tabulary,overpic,xcolor}
\definecolor{mygray}{gray}{.92}
\definecolor{myred}{rgb}{1,0.5,0.5}
\definecolor{myred2}{rgb}{0.75,0,0}
\usepackage{pifont}
\newcommand{\xmark}{\ding{55}}
\newcommand{\myparagraph}[1]{{\vspace{0.5em} \noindent \bf #1}}
\newlength\savewidth\newcommand\shline{\noalign{\global\savewidth\arrayrulewidth
		\global\arrayrulewidth 1pt}\hline\noalign{\global\arrayrulewidth\savewidth}}
\newcolumntype{x}[1]{>{\centering\arraybackslash}p{#1pt}}
\newcommand{\tablestyle}[2]{\setlength{\tabcolsep}{#1}\renewcommand{\arraystretch}{#2}\centering\footnotesize}

\usepackage[utf8]{inputenc}
\newcommand{\authorskip}{\hspace{12mm}}
\usepackage{caption}
\usepackage{cuted}

\usepackage[pagebackref,breaklinks,colorlinks]{hyperref}

\usepackage[capitalize]{cleveref}
\crefname{section}{§}{§§}
\Crefname{section}{§}{§§}

\def\cvprPaperID{2216} 
\def\confName{CVPR}
\def\confYear{2022}

\begin{document}

\title{Efficient Video Instance Segmentation via Tracklet Query and Proposal}

\author{
 Jialian Wu$^{1}$ \authorskip Sudhir Yarram$^{1}$ \authorskip Hui Liang$^{2}$ \authorskip
 Tian Lan$^{2}$ \\ Junsong Yuan$^{1}$ \authorskip Jayan Eledath$^{2}$ \authorskip G\'{e}rard Medioni$^{2}$\\[3mm]
 $^1$State University of New York at Buffalo ~~~~~~~~~~~~~~ $^2$Amazon \\
 {\small \url{https://jialianwu.com/projects/EfficientVIS.html}}
}
\maketitle
	
\begin{abstract} 
Video Instance Segmentation (VIS) aims to simultaneously classify, segment, and track multiple object instances in videos. Recent clip-level VIS takes a short video clip as input each time showing stronger performance than frame-level VIS (tracking-by-segmentation), as more temporal context from multiple frames is utilized. Yet, most clip-level methods are neither end-to-end learnable nor real-time. These limitations are addressed by the recent VIS transformer (VisTR)~\cite{vistr} which performs VIS end-to-end within a clip. However, VisTR suffers from long training time due to its frame-wise dense attention. In addition, VisTR is not fully end-to-end learnable in multiple video clips as it requires a hand-crafted data association to link instance tracklets between successive clips. This paper proposes EfficientVIS, a fully end-to-end framework with efficient training and inference. At the core are tracklet query and tracklet proposal that associate and segment regions-of-interest (RoIs) across space and time by an iterative query-video interaction. We further propose a correspondence learning that makes tracklets linking between clips end-to-end learnable. Compared to VisTR,  EfficientVIS requires $15\times$ fewer training epochs while achieving state-of-the-art accuracy on the YouTube-VIS benchmark. Meanwhile, our method enables whole video instance segmentation in a single end-to-end pass without data association at all.

\end{abstract}
\vspace{-4mm}
\section{Introduction}
\label{sec:intro}
\begin{figure}
	\centering
	\includegraphics[width=1\linewidth]{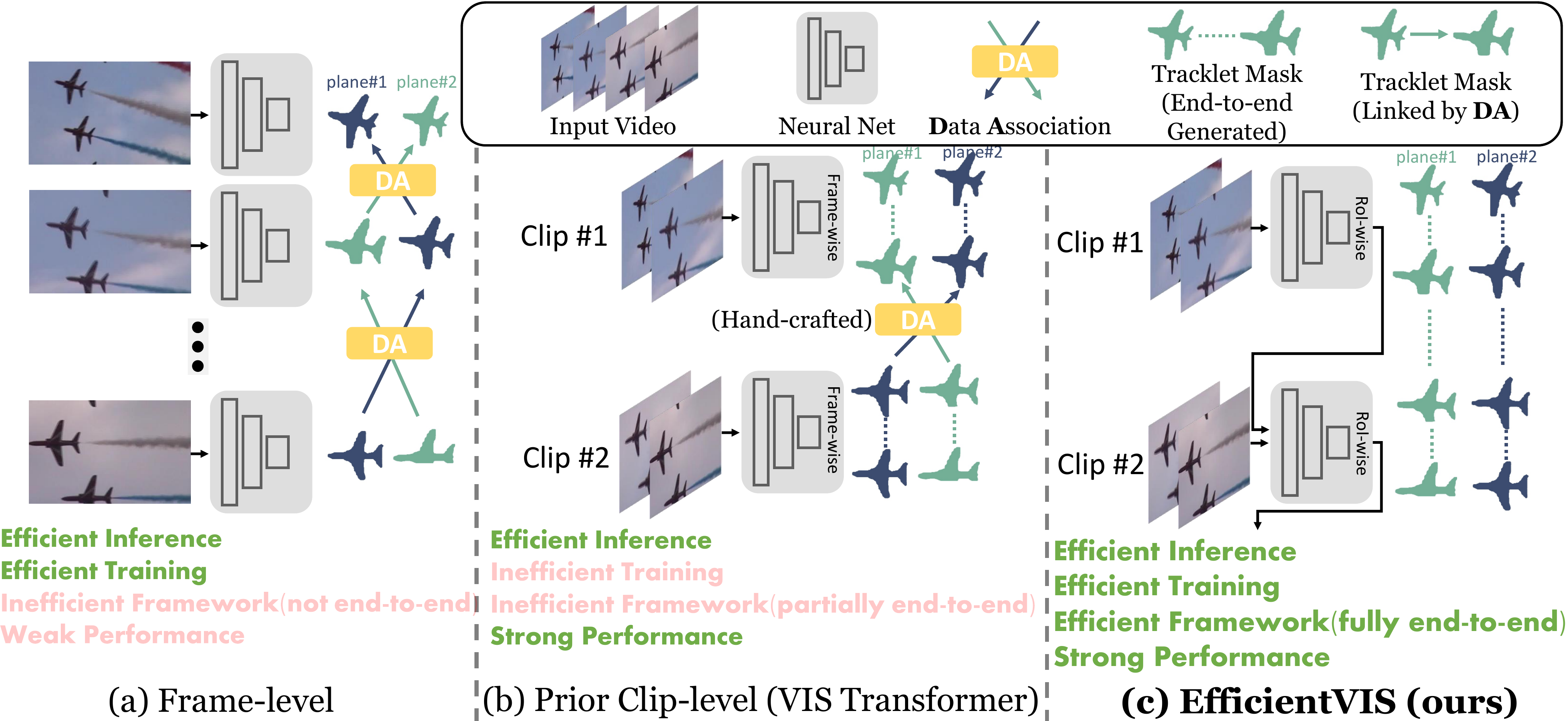}
	\vspace{-5mm}
	\caption{\textbf{Overview of real-time VIS pipelines.} Compared to (a) or (b), our method features \textbf{1) Efficient training} thanks to the RoI-wise design; \textbf{2) Efficient framework} that achieves whole video instance segmentation in a single end-to-end pass without any data association and post-processing; \textbf{3) Strong performance} thanks to the clip-by-clip processing and fully end-to-end paradigm.}
	\label{fig:fig1}
	\vspace{-2mm}
\end{figure}

Video Instance Segmentation (VIS) is a challenging video task recently introduced in~\cite{yang2019video}. It aims to predict a tracklet segmentation mask with a class label for every appeared object instance in a video as illustrated in Fig.~\ref{fig:fig1}. Existing methods typically solve the VIS problem at either frame-level or clip-level. The frame-level methods~\cite{Yang_2021_ICCV,QueryInst,li2021spatial,liu2021sg,wu2021track,wang2021end2} follow a tracking-by-segmentation paradigm, which first performs image instance segmentation and then links the current masks with history tracklets via data association as shown in Fig.~\ref{fig:fig1} (a). This paradigm typically requires complex data association algorithms and exploits limited temporal context, making it susceptible to object occlusions. By contrast, the clip-level methods~\cite{athar2020stem,bertasius2020classifying,lin2021video,vistr} jointly perform segmentation and tracking clip-by-clip. Within each clip, object information is propagated back and forth. Such a paradigm usually performs stronger than the frame-level methods thanks to larger temporal receptive field. However, most clip-level methods are not end-to-end and require elaborated inference that causes slow speed. These issues are addressed by the recent VIS transformer (VisTR)~\cite{vistr} which extends image object detector DETR~\cite{carion2020end} to the VIS task. VisTR generates VIS predictions within each video clip in one end-to-end pass, which greatly simplifies the clip-level paradigm and makes VIS within a clip end-to-end trainable.
	
However, the VIS transformer is confronted with two issues: \textbf{(i)} The convergence speed of VisTR is slow since the frame-wise dense attention weights in the transformer need long training epochs to search for properly attended regions among all video frame pixels. It is difficult for researchers to experiment with this algorithm as it requires long development cycles. \textbf{(ii)} When a video is too long to fit into GPU memory in one forward-pass, it has to be divided into multiple successive clips for sequential processing. In this case, VisTR becomes partially end-to-end. As shown in Fig.~\ref{fig:fig1} (b), VisTR requires a hand-crafted data association to link tracklets between clips. Such a hand-crafted scheme is not only more complicated but also less effective than an end-to-end learnable association as evidenced in Table~\ref{tab:ablation4}. 

In this paper, we propose a new clip-level VIS framework, coined as EfficientVIS. EfficientVIS delivers a fully end-to-end framework that can achieve fast convergence and inference with strong performance. Our work is inspired by the recent success of the query-based R-CNN architecture~\cite{sun2021sparse,QueryInst} in image object detection. EfficientVIS extends the spirit to the video domain to model the complex space-time interactions. Specifically, EfficientVIS uses \emph{tracklet query} paired with \emph{tracklet proposal} to represent an individual object instance in video. Tracklet query is latent embeddings that encode appearance information for a target instance, while tracklet proposal is a space-time tube that locates the target in the video. Our tracklet query collects the target information by interacting with video in a \emph{clip-by-clip} fashion through a designed \emph{temporal dynamic convolution}. This way enriches temporal object context that is an important cue for handling object occlusion and motion blur in videos. We also design a \emph{factorised temporo-spatial self-attention} allowing tracklet queries to exchange information over not only space but also time. It enables one query to correlate a target across multiple frames so as to end-to-end generate target tracklet mask as a whole in a video clip. Compared to prior works, EfficientVIS enjoys three remarkable properties:

\emph{(i) Fast convergence:} In EfficientVIS, tracklet queries interact with video CNN features only in the region of the space-time RoIs defined by the tracklet proposal. This is different from VisTR~\cite{vistr} that is interacting with all video pixels on the transformer~\cite{vaswani2017attention} encoded features using dense attention. Our RoI-wise design drastically reduces video redundancies and therefore allows EfficientVIS for faster convergence than transformer as shown in Table~\ref{tab:ablation8}.

\emph{(ii) Fully end-to-end learnable framework:} EfficientVIS goes beyond a short clip and is fully end-to-end learnable over the whole video. When there are multiple successive clips, one needs to link instance tracklets between clips. In contrast to prior clip-level works~\cite{vistr,bertasius2020classifying,pang2020tubetk,Wang_2020_CVPR} that manually stitch the tracklets, we design a \emph{correspondence learning} that enables tracklet query to be shared among clips for \emph{seamlessly} associating a same instance. In other words, the tracklet query output from one clip is enabled to be fed into the next clip to associate and segment the same instance. Meanwhile, the query is dynamically updated in terms of the next clip content so as to achieve a continuous tracking for the future. Such a scheme makes EfficientVIS fully end-to-end, without any explicit data association for either inner-clip or inter-clip tracking as shown in Fig.~\ref{fig:fig1} (c).

\emph{(iii) Tracking in low frame rate videos:} Tracking object instances that have dramatic movements is a great challenge for motion-based trackers~\cite{pang2020tubetk,Wang_2020_CVPR,zhou2020tracking}, as they suppose instances move smoothly over time. In contrast, our method retrieves a target instance in a frame conditioned on its query representation regardless of where it is in nearby frames. Thus, EfficientVIS is robust to dramatic object movements and can track instances in low frame rate videos as shown in Fig.~\ref{fig:fig5} and Table~\ref{tab:ablation7}.

We summarize our major contributions as follows:
\begin{itemize}
\vspace{-2mm}
	\item \textbf{EfficientVIS is the first RoI-wise clip-level VIS framework that runs in real-time.}
	The RoI-wise design enables a fast convergence by drastically reducing video redundancies. Fully end-to-end learnable tracking and rich temporal context of the clip-by-clip workflow together bring a strong performance. EfficientVIS ResNet-50 achieves 37.9 AP on Youtube-VIS~\cite{yang2019video} in 36 FPS by training 33 epochs, which is 15$\times$ training epochs fewer and 2.3 AP higher than VIS transformer.
		
	\vspace{-3mm}
	\item \textbf{EfficientVIS is the first fully end-to-end neural net for VIS.} Given a video as input despite its length, EfficientVIS directly produces VIS predictions without any data association or post-processing. We will demonstrate by diagnostic experiments that this fully end-to-end paradigm is not only simpler but also more effective than the previous partially/non end-to-end frameworks.
\end{itemize}
	
\section{Related Works}	
\vspace{-3mm}
\myparagraph{Frame-level VIS:} Most video instance segmentation methods work at the frame-level fashion, \emph{a.k.a.} tracking-by-segmentation~\cite{Yang_2021_ICCV,QueryInst,li2021spatial,liu2021sg,wu2021track,cao2020sipmask,wang2021end2,fu2020compfeat,qin2021learning}. This paradigm produces instance segmentation frame-by-frame and achieves tracking by linking the current instance mask to the history tracklet. So it requires an explicit data association algorithm in either a hand-crafted or trainable way. Some attempts~\cite{wu2021track,Yang_2021_ICCV,fu2020compfeat,li2021spatial} strive to exploit temporal context to improve VIS, yet their temporal context is limited to one or only a few frames ignoring fertile resources of videos. In contrast to the above methods, our method does not require data association algorithm at all and we take advantage of richer temporal context by performing VIS clip-by-clip.
	
\begin{figure*}[h]
	\centering
	\includegraphics[width=1\linewidth]{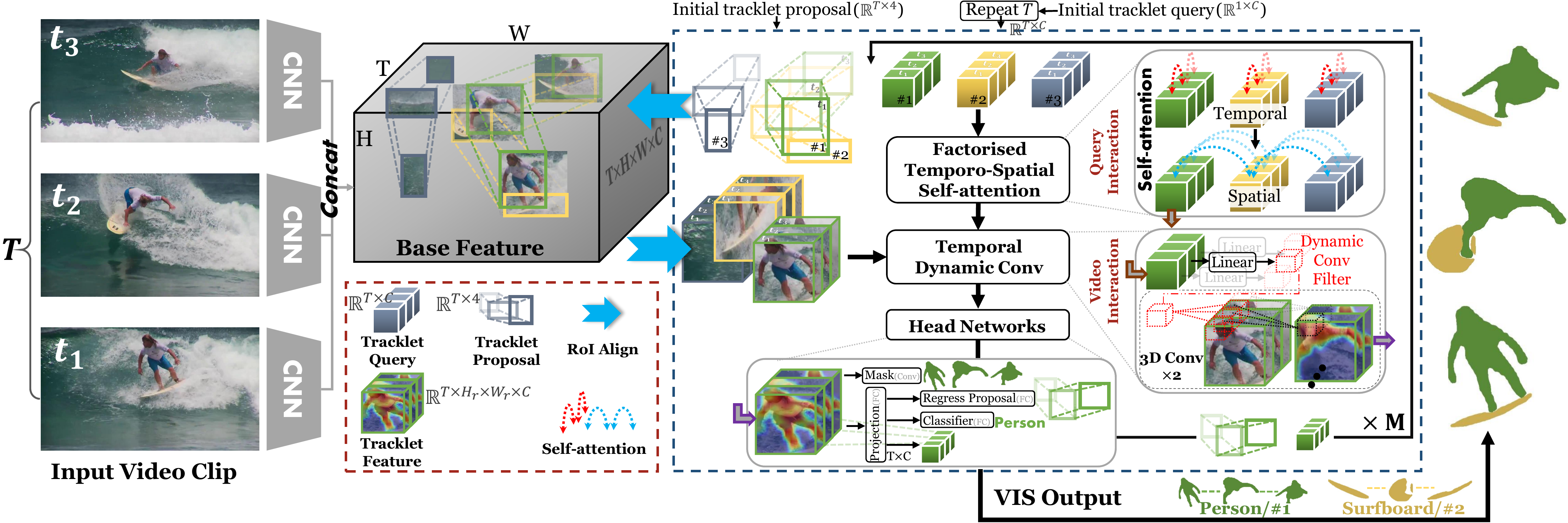}
	\vspace{-6mm}
	\caption{\textbf{EfficientVIS architecture.} EfficientVIS performs VIS clip-by-clip where the above figure illustrates how it works in one clip.}
	\label{fig:overview}
	\vspace{-4mm}
\end{figure*}

\vspace{-1mm}
\myparagraph{Clip-level VIS:} Recent clip-level works~\cite{athar2020stem,bertasius2020classifying,lin2021video,vistr} demonstrate promising VIS accuracy by exploiting rich temporal knowledge from multiple frames of a video clip. This paradigm propagates object information back and forth in a clip, which can well handle object occlusion and motion blur. However, most clip-level methods have a complex inference process (\eg extra mask refinement, box ensemble, etc) that makes them neither end-to-end learnable nor real-time in inference. To address these limitations, the VIS transformer (VisTR)~\cite{vistr} extends DETR~\cite{carion2020end} from images to videos, achieving efficient and end-to-end VIS within each clip. Nevertheless, VisTR suffers from slow convergence due to the frame-wise dense attention. Besides, if there are two or more video clips, VisTR is not fully end-to-end and requires a hand-crafted data association to link tracklets between clips. In contrast to the above methods, EfficientVIS is fully end-to-end learnable without the need of data association for either inner- or inter-clips. Moreover, EfficientVIS performs VIS in an RoI-wise manner through an efficient query-video interaction, enabling fast convergence and inference. 
	
\section{EfficientVIS}
\vspace{-1mm}
EfficientVIS aims to deliver four key features: \textbf{(i)} \emph{Fast training and inference}; \textbf{(ii)} \emph{Fully end-to-end learnable framework} without any data association/post-processing; \textbf{(iii)} \emph{Tracking in low frame rate videos}; \textbf{(iv)} \emph{Strong performance}. EfficientVIS performs VIS clip-by-clip, where we present the main VIS architecture for inner-clip in Sec.~\ref{subsec:Architecture} and the inter-clip tracklets correspondence learning in Sec.~\ref{subsec:Correspondence_learning}.
\vspace{-1mm}
\subsection{Main Architecture}
\label{subsec:Architecture}
EfficientVIS is an RoI-wise clip-level VIS framework. In each forward-pass, it takes as input a video clip $\{\bm{I}_{t}\}^{T}_{t=1}$ and directly yields the VIS predictions, \ie tracklet masks $\{\bm{m}_{i}\}^{N}_{i=1}$ along with tracklet classifications, where $\bm{I}_{t}\in \mathbb{R}^{H_{I} \times W_{I} \times 3}$ and $\bm{m}_{i}\in \mathbb{R}^{T\times H_{I} \times W_{I} \times 1}$. $T$ is the number of frames in the clip, and $N$ is the number of unique object instances. $H_{I} \times W_{I}$ is the frame spatial size. As detailed in Fig.~\ref{fig:overview}, EfficientVIS starts with a CNN backbone that extracts a video base feature. Afterward, EfficientVIS iterates over the following query-video interaction $M$ times: 1) Query interaction: tracklet queries self-interact and communicate in space and time by our \emph{factorised temporo-spatial self-attention} so that a tracklet query consistently associates the same instance over time and different tracklet queries track different instances; 2) Video interaction: tracklet queries interact with video feature to collect the target instance information by our \emph{temporal dynamic convolution}. At the end of each interaction, a couple of head networks are applied to update tracklet queries, proposals, masks, and classifications. Finally, EfficientVIS takes tracklet masks and classifications output from the last iteration as the VIS results.

\vspace{-2mm}
\myparagraph{Base Feature:} We apply a standard CNN backbone frame-by-frame to the input video clip. The extracted features are then concatenated along the time dimension to form the base feature $\bm{f}\in \mathbb{R}^{T\times H \times W \times C}$, where $C$ is the number of channels and  $H \times W$ is the spatial size of feature map. The base feature contains full semantic information of the clip.

\vspace{-2mm}
\myparagraph{Tracklet Query and Proposal:} We use tracklet queries $\{\bm{q}_{i}\}^{N}_{i=1}$ paired with tracklet proposals $\{\bm{b}_{i}\}^{N}_{i=1}$ to represent every unique object instance throughout a video. Tracklet query $\bm{q}_{i}$ is embedding vectors that encode exclusive appearance information for the $i$-th instance, indicating the instance identity. Tracklet proposal is a space-time box tube $\bm{b}_{i} \in \mathbb{R}^{T\times 4}$, which tracks and locates the $i$-th instance across the entire video clip. $\bm{b}_{i}^{t}\in \mathbb{R}^{4}$ determines the four corners of the box at time $t$. Each tracklet query has $T$ embeddings, \ie $\bm{q}_{i} \in \mathbb{R}^{T \times C}$, where $C$ is the channel number of an embedding. In this way, we enforce different embeddings to focus on different frames. This is more effective than using only one embedding as evidenced in Table~\ref{tab:ablation6}, because one instance at different time could have significant appearance changes. It is harsh to enforce a single embedding to encode various appearance patterns of an instance.

\vspace{-2mm}
\myparagraph{Factorised Temporo-Spatial Self-Attention (FTSA):} The goal of FTSA is to achieve a query-target correspondence that makes each query associate to a unique target instance in the video. We take advantage of the idea of factorised temporo-spatial self-attention in video transformer~\cite{gberta_2021_ICML,Arnab_2021_ICCV} which is originally used for encoding video backbone features. Here, we exploit it on instance tracklet queries to achieve an instance communication. Concretely, we separately perform temporal and spatial Multi-Head Self-Attention (MSA)~\cite{vaswani2017attention} one after the other on tracklet queries as shown in Fig.~\ref{fig:overview}. Let $\bm{q}^{t}_{i}\in \mathbb{R}^{1 \times C}$ denote the $t$-th embedding in the $i$-th instance tracklet query. The temporal self-attention is computed within the same tracklet query, across the $T$ embedding pairs as:
\vspace{-2mm}
\begin{equation}
\vspace{-2mm}
\label{eqn:temporal_self_attention}
\{\bm{q}^{t}_{i}\}^{T}_{t=1} \leftarrow \text{MSA}(\{\bm{q}^{t}_{i}\}^{T}_{t=1}), \quad i=1,...,N.
\end{equation}
The temporal self-attention allows the embeddings of one instance query from different frames to exchange the target instance information, such that the embeddings can jointly associate the same instance over time. Therefore, instance association/tracking in each clip is \textbf{implicitly} achieved and end-to-end \textbf{learned} by our temporal self-attention. The spatial self-attention is then computed at the same frame, across the $N$ embedding pairs from different tracklet queries as:
\vspace{-2mm}
\begin{equation}
\vspace{-2mm}
\label{eqn:spatial_self_attention}
\{\bm{q}^{t}_{i}\}^{N}_{i=1} \leftarrow \text{MSA}(\{\bm{q}^{t}_{i}\}^{N}_{i=1}), \quad t=1,...,T.
\end{equation}
The spatial self-attention allows each query to acquire object context from other queries in each frame, reasoning about the relations among different object instances. We observe in experiments that the order of temporal and spatial self-attention does not cause a noticeable performance difference. 
	
Joint temporo-spatial self-attention that computes MSA across all $T\times N$ embedding pairs for once is another alternative for query communication. Compared to this scheme, our FTSA saves more computation and is more effective as we will demonstrate in Table~\ref{tab:ablation1}.

\vspace{-1mm}
\myparagraph{Temporal Dynamic Convolution (TDC):} The aim of TDC is to collect the target instance information from the video clip. Specifically, we generate a dynamic convolutional filter~\cite{jia2016dynamic,tian2020conditional} conditioned on tracklet query embedding. We then use it to perform convolution on an RoI region of the base feature specified by the corresponding tracklet proposal. Different from the still-image dynamic convolution~\cite{tian2020conditional,sun2021sparse,QueryInst}, we perform a 3D dynamic convolution in order to collect temporal instance context from nearby frames as well. Let $\bm{w}^{t}_{i}$ denote the dynamic convolutional filter generated from tracklet query embedding $\bm{q}^{t}_{i}$. The TDC is computed as:
\vspace{-2mm}
\begin{equation}
\vspace{-2mm}
\label{eqn:tdn}
\bm{o}^{t}_{i} = \sum_{t^{\prime}=t-1}^{t+1}\bm{a}_{i,(t,t^{\prime})} \circ \text{conv2d}(\bm{w}^{t}_{i}, \phi(\bm{f}^{t^{\prime}}, \bm{b}^{t^{\prime}}_{i})).
\end{equation}
$\phi$ denotes RoI align~\cite{he2017mask} whose output spatial size is $H_r\times W_r$ and $\circ$ is element-wise product. $\bm{a}_{i,(t,t^{\prime})}\in \mathbb{R}^{H_r\times W_r}$ is a simple adaptive weight that measures the similarity of conv2d($\cdot$) outputs between time step $t$ and $t^{\prime}$. Similar to~\cite{wu2020temporal,zhu2017flow}, the adaptive weight is calculated by cosine similarity and softmax such that $\sum_{t^{\prime}=t-1}^{t+1}a_{i,(t,t^{\prime})}^{x,y}=1$. $\{\bm{o}^{t}_{i}\}_{t=1}^{T}\in \mathbb{R}^{T \times H_r\times W_r \times C}$ is the \emph{tracklet feature} of the $i$-th instance. Eq.~\ref{eqn:tdn} differs from regular 3D convolution, where we share the same filter over the time dimension. 
	
Since the dynamic filter $\bm{w}_{i}$ is generated from $\bm{q}_{i}$, $\bm{w}_{i}$ has distinct cues and appearance semantics of the $i$-th instance. The TDC exploits $\bm{w}_{i}$ to exclusively collect the $i$-th instance information from the video base feature yet filter out irrelevant information. Therefore, tracklet feature $\bm{o}_{i}$ shall be highly activated by $\bm{w}_{i}$ in the region where the $i$-th instance appears but will be suppressed in the region of uncorrelated instances or background clutter as evidenced in Fig.~\ref{fig:fig4}. 

\vspace{-1mm}
\myparagraph{Head Networks:} For each instance $i$, we employ light-weight head networks on its tracklet feature $\bm{o}_{i}$ to output its VIS predictions and renewed tracklet query and proposal. Since tracklet feature $\bm{o}_{i}$ exclusively carries the $i$-th instance information across the clip, these outputs can be readily derived by applying regular convolutions or fully connected layers (FCs). Specifically, we apply convolutions to segment tracklet mask, and FCs to classify tracklet, regress tracklet proposal, and update tracklet query. The updated query and proposal are fed into the next iteration. The head network architectures are similar to~\cite{he2017mask,sun2021sparse,QueryInst}. In a nutshell, our tracklet (proposal) for a video clip is directly generated as a whole determined by tracklet query. So we do not need explicit data association to link instances/boxes across frames in a clip as frame-level VIS methods~\cite{Yang_2021_ICCV,QueryInst,li2021spatial,liu2021sg,wang2021end2,fu2020compfeat,qin2021learning}.

\vspace{-1mm}
\myparagraph{Initial Tracklet Query and Proposal Details:} As~\cite{sun2021sparse,QueryInst}, the initial tracklet query and proposal at the input of the first iteration are model parameters learned from training. Each initial query has one embedding, \ie $\bm{q}^{\star}_{i} \in \mathbb{R}^{1 \times C}$. As shown in Fig.~\ref{fig:overview}, we repeat it $T$ times over the time dimension before starting the first iteration\footnote{We repeat the parameter $\bm{q}^{\star}$ $T$ times rather than directly setting the parameter to be $\bm{q}^{\star} \in \mathbb{R}^{T \times C}$, because it gives an initial prior that all the $T$ embeddings shall belong to a same instance.}. Besides, we find in experiments that all the learned initial proposals tend to be the frame size which covers the whole scene. We think this is reasonable because it is easier for later stages to regress the proposal to the target region by first taking a glance at the whole image.
	
\subsection{Training}
\label{subsec:Training}
For each video clip, EfficientVIS is trained by the one-to-one matching loss~\cite{carion2020end,sun2021sparse,QueryInst,vistr}, which first obtains a one-to-one assignment between predictions and ground truths by bipartite matching and then computes the training loss in terms of the assignment.  Let $\{y_{i}\}_{i=1}^{N}$ denote the tracklet predictions and $\{\hat{y}_{j}\}_{j=1}^{G}$ denote the tracklet ground truths, where $G$ ($\leq N$) is the total number of ground truths. We denote $\sigma$ as a $G$-elements permutation of $\{1,...,N\}$. The bipartite matching between $y$ and $\hat{y}$ is found by searching for a $\sigma$ with the lowest cost as:
\vspace{-3mm}
\begin{equation}
\vspace{-2mm}
\hat{\sigma}=\underset{\sigma}{\arg \min } \sum_{j=1}^{G} \mathcal{L}_{\text{match}}\left(y_{\sigma(j)}, \hat{y}_{j}\right).
\end{equation}
The optimal assignment $\hat{\sigma}$ is obtained by the Hungarian algorithm. Each instance tracklet has three types of annotation $\hat{y}_{j}=(\hat{\bm{c}}_{j},\hat{\bm{b}}_{j},\hat{\bm{m}}_{j})$, where $\hat{\bm{c}}_{j} \in \mathbb{R}^{T}$ is the class label including background class $\emptyset$. $\hat{c}^{t}_{j}=\emptyset$ indicates the instance disappears at time step $t$. Let $p_{\sigma(j)}(\hat{\bm{c}}_{j})$ be the predicted classification score of the $\sigma(j)$-th tracklet for the class $\hat{\bm{c}}_{j}$. The matching cost is computed by $\mathcal{L}_{\text{match}}=\sum_{t=1}^{T}-p_{\sigma(j)}(\hat{c}^{t}_{j})+\mathds{1}_{\{\hat{c}^{t}_{j}\neq \emptyset\}}(\mathcal{L}_{\text{box}}(\bm{b}_{\sigma(j)}^{t}, \hat{\bm{b}}_{j}^{t})-\mathcal{L}_{\text{maskIoU}}(\bm{m}_{\sigma(j)}^{t}, \hat{\bm{m}}_{j}^{t}))$, where $\bm{b}$ and $\bm{m}$ are the predicted tracklet proposal and mask respectively. $\mathcal{L}_{\text{box}}$ is calculated by box IoU and L1 distance as~\cite{carion2020end}. After finding the optimal assignment $\hat{\sigma}$ that one-to-one matches between tracklet predictions and ground truths, the training loss for the video clip is computed as:
\vspace{-3mm}
\begin{equation}
\vspace{-2mm}
\label{eqn:overall_loss}
\begin{split}
&\mathcal{L}_{\text{clip}}(\hat{\sigma})=\sum_{j=1}^{G}\sum_{t=1}^{T}[-p_{\hat{\sigma}(j)}(\hat{c}^{t}_{j})+\mathcal{L}_{\text{CE}}(\bm{m}_{\hat{\sigma}(j)}^{t}, \hat{\bm{m}}_{j}^{t})\\
&+\mathds{1}_{\{\hat{c}^{t}_{j}\neq \emptyset\}}(\mathcal{L}_{\text{box}}(\bm{b}_{\hat{\sigma}(j)}^{t}, \hat{\bm{b}}_{j}^{t})+\mathcal{L}_{\text{dice}}(\bm{m}_{\hat{\sigma}(j)}^{t}, \hat{\bm{m}}_{j}^{t}))],
\end{split}
\end{equation}
where $\mathcal{L}_{\text{CE}}$ is the binary cross-entropy loss and $\mathcal{L}_{\text{dice}}$ is the dice loss~\cite{milletari2016v}. All the losses and balancing weights are similar to~\cite{carion2020end,vistr,QueryInst,sun2021sparse}, except that we add a $\mathcal{L}_{\text{CE}}$ to enforce predicted mask to be zero if the instance disappears. Eq.~\ref{eqn:overall_loss} is averaged by the total number of entries. For those unmatched tracklet predictions, we only impose classification loss to supervise them to be background class.
	
\begin{figure}
	\centering
	\includegraphics[width=1\linewidth]{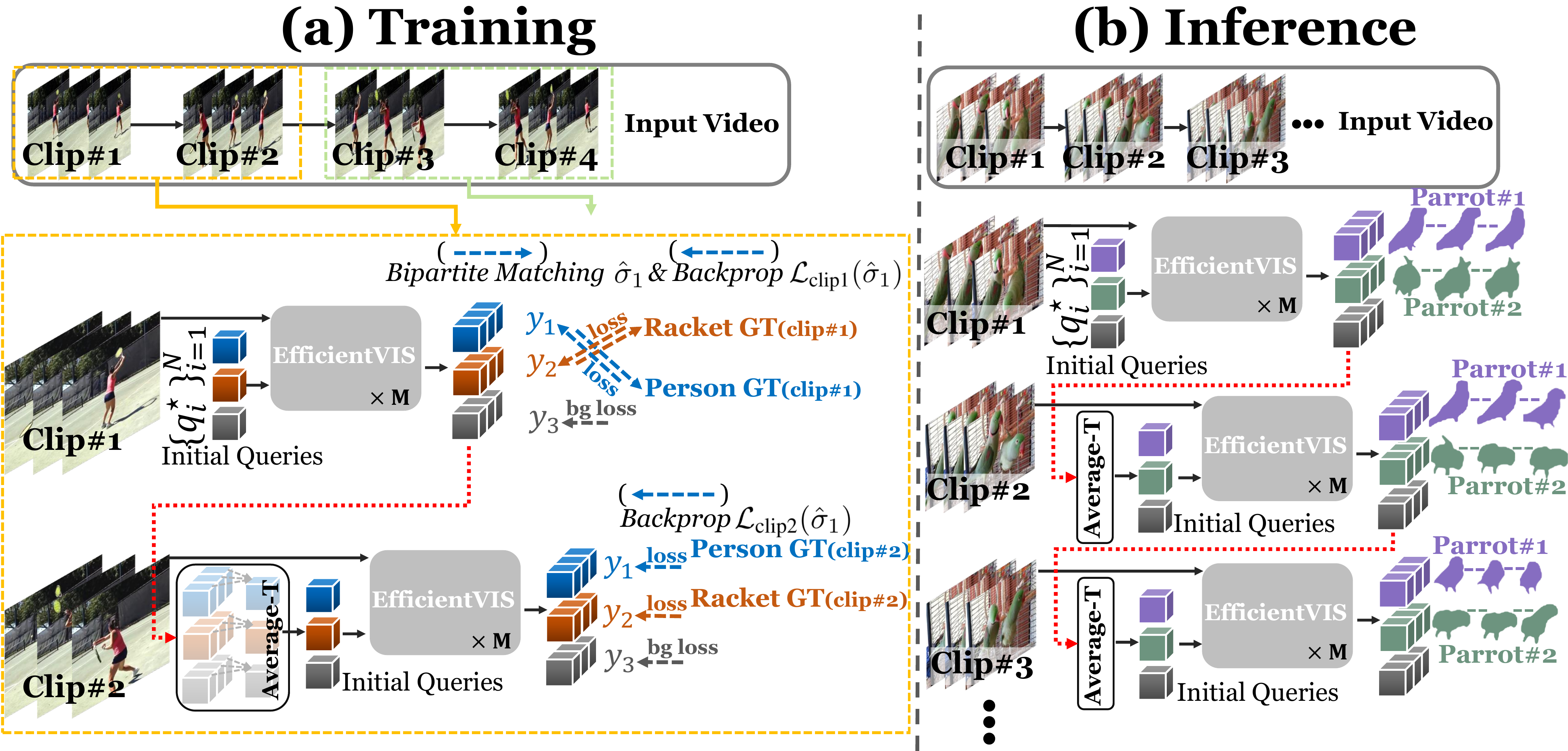}
	\vspace{-7mm}
	\caption{\textbf{(a) Correspondence learning}. \textbf{(b) Fully end-to-end inference in multiple clips}.}
	\label{fig:fig3}
	\vspace{-3mm}
\end{figure}
	
\subsection{Correspondence Learning}
\label{subsec:Correspondence_learning}
If a long video cannot fit into GPU memory in one forward-pass, it needs to be split into multiple successive clips for sequential processing. To stitch tracklets between two clips, prior works~\cite{vistr,bertasius2020classifying,pang2020tubetk,Wang_2020_CVPR} turn to hand-crafted data association. However, human-designed rules are cumbersome and not end-to-end learnable. Therefore, we design a correspondence learning that makes the tracklet linking between clips end-to-end learnable without data association, which is not only simpler but also more effective.

\vspace{-2mm}
\myparagraph{Correspondence learning:} As shown in Fig.~\ref{fig:fig3}, every two clips from the input video are paired\footnote{The order of the two clips in a pair is randomized.} in training. For the first clip of a pair, the initial tracklet queries are the model parameters $\bm{q}^{\star}$ as described in Sec.~\ref{subsec:Architecture}, and the training procedure is the same as Sec.~\ref{subsec:Training}, \ie finding the optimal assignment $\hat{\sigma}$ and backpropagating $\mathcal{L}_{\text{clip}}(\hat{\sigma})$. For the initial queries of the second clip, we use the output tracklet queries from the first clip averaged along the time dimension. The average operation can extract a comprehensive abstract for the instance representation. As for the assignment of the second clip, we do not perform bipartite matching. Instead, we use the same assignment $\hat{\sigma}$ that is already obtained by the first clip to train the second clip. Such a training manner enforces the tracklet queries output from one clip to segment and correspond to the same object instances in the next clip. In this way, tracklet query is able to serve as a generic video-level representation to \emph{seamlessly} associate and correlate an instance across a whole video instead of only a clip. 

\myparagraph{Inference:} EfficientVIS simply takes output queries from the last clip as the initial queries to correlate and segment the same instances in the current clip without data association. Meanwhile, the queries are continuously updated by our query-video interaction to collect the current clip information so as to achieve a continuous tracking for the future. If a tracklet query has been classified as background for many clips, one may re-initialize it with $\bm{q}^{\star}$ in order to allocate slots. Since instance positions vary in different clips, we use the same initial proposals for every clip, \ie the trained model parameters rather than the last clip output proposals.

\section{Experiments}
\vspace{-1mm}
\subsection{Experimental Setup}
\vspace{-3mm}
\myparagraph{Datasets:} Our experiments are conducted on the YouTube-VIS~\cite{yang2019video} benchmark in two versions. YouTube-VIS 2019 is the first dataset for video instance segmentation, which contains 2,883 videos labeled at 6 FPS, 131K instance masks, and 40 object classes. YouTube-VIS 2021 is an increased version that contains 3859 videos.

\vspace{-1mm}	
\myparagraph{Evaluation Metric:} The video-level average precision (AP) and average recall (AR) are the metrics. Different from the image domain, the video intersection over union (IoU) is computed between the predicted tracklet mask and ground truth tracklet mask. So, in order to achieve high performance, a model needs to not only correctly classify and segment the target instance but also accurately track the target over time. 
	
\begin{table*}[t]
	\small
	\begin{subtable}{0.5\linewidth}
		\centering
		\captionsetup{width=0.9\linewidth}
		\begin{tabular}{c|c|cccc} 	
		\rowcolor{mygray}
		Self-attention&AP&AP$_{50}$&AP$_{75}$&AR$_{1}$&AR$_{10}$\\
		\shline
		Spatial (S)&32.8&56.3&34.9&36.1&42.2\\
		Temporal (T)&30.4&49.0&32.1&34.2&39.1\\
		Joint T-S&33.7&56.8&36.1&36.6&44.6\\
		Factorised T-S (FTSA)&\bf 37.0&59.6&40.0&39.3&46.3\\
		\end{tabular}
		\caption{\textbf{Self-attention schemes.} We perform different multi-head self-attention schemes on tracklet queries.}
		\label{tab:ablation1}
	\end{subtable}
	\begin{subtable}{0.5\linewidth}
		\centering
		\captionsetup{width=0.9\linewidth}
		\begin{tabular}{c|c|cccc} 	
		\rowcolor{mygray}
		Dynamic Conv&AP&AP$_{50}$&AP$_{75}$&AR$_{1}$&AR$_{10}$\\
		\shline
		Still-image&36.0&59.5&39.5&38.4&45.3\\
		Temporal &\bf 37.0&59.6&40.0&39.3&46.3\\
		\end{tabular}
		\caption{\textbf{Still-image \emph{vs.} Temporal - dynamic convolution}. Temporal dynamic convolution is more effective by taking into account temporal object context from nearby frames.}
		\label{tab:ablation2}
	\end{subtable}
	\vspace{-2mm}
		
	\begin{subtable}{0.5\linewidth}
		\centering
		\vspace{-3mm}
		\captionsetup{width=0.9\linewidth}
		\begin{tabular}{c|c|cccc}	
		\rowcolor{mygray}
		Length &AP&AP$_{50}$&AP$_{75}$&AR$_{1}$&AR$_{10}$ \\
		\shline
		$T=9$ &35.3&57.6&38.5&37.7&43.5\\
		$T=18$&36.4&58.1&39.5&39.1&46.8\\
		$T=36$&\bf37.0&59.6&40.0&39.3&46.3\\	
		\end{tabular}
		\caption{\textbf{Video clip length $T$.} We experiment with different number of frames for each video clip. Larger temporal receptive filed provides richer temporal context and therefore yields better performance.}
		\label{tab:ablation3}
	\end{subtable}
	\begin{subtable}{0.5\linewidth}
		\centering
		\captionsetup{width=0.9\linewidth}
		\tablestyle{4pt}{1.05}
		\begin{tabular}{c|c|l|cccc}
		\rowcolor{mygray}
		\multicolumn{2}{c|}{Scheme} &AP&AP$_{50}$&AP$_{75}$&AR$_{1}$&AR$_{10}$\\
		\shline
		\multirow{2}{*}{$T=9$}
		&Hand-craft&33.7\textcolor{myred}{(\textbf{$-$1.6})}&55.5&36.4&33.9&40.3\\
		&Fully e2e (ours)&\bf 35.3&57.6&38.5&37.7&43.5\\
		\hline
		\multirow{3}{*}{$T=18$}
		&VisTR~\cite{vistr}&29.7\textcolor{myred}{(\textbf{$-$6.7})}&50.4 &31.1&29.5&34.4\\
		&Hand-craft&34.6\textcolor{myred}{(\textbf{$-$1.8})}&55.3&37.4&36.6&44.6\\
		&Fully e2e (ours)&\bf 36.4&58.1&39.5&39.1&46.8\\
		\end{tabular}
		\caption{\textbf{Fully end-to-end (e2e) \emph{vs.} Partially e2e}. Both ``Hand-craft" and VisTR are partially e2e, where tracklet association within each clip is e2e but that between clips requires a hand-crafted linking. For ``Hand-craft", we report the best results by varying matching score thresholds.}
		\label{tab:ablation4}
	\end{subtable}
		
	\begin{subtable}{0.5\linewidth}
		\centering
		\captionsetup{width=0.9\linewidth}
		\vspace{-8mm}
		\begin{tabular}{c|c|cccc}
		\rowcolor{mygray}
		Scheme&AP&AP$_{50}$&AP$_{75}$&AR$_{1}$&AR$_{10}$\\
		\shline
		w/o CL&7.4&19.0&5.9&9.8&13.5\\
		w/ CL&\bf 35.3&57.6&38.5&37.7&43.5\\
		\end{tabular}
		\caption{\textbf{Correspondence learning (CL).} We train EfficientVIS with and without CL. After training, we test both using our fully end-to-end inference paradigm. $T=9$ in this study.}
		\label{tab:ablation5}
	\end{subtable}
	\begin{subtable}{0.5\linewidth}
		\centering
		\captionsetup{width=0.9\linewidth}
		\begin{tabular}{c|c|cccc}
		\rowcolor{mygray}
		Query&AP&AP$_{50}$&AP$_{75}$&AR$_{1}$&AR$_{10}$\\
		\shline
		Time shared&35.5&57.1&38.5&38.7&43.9\\
		Time disentangled&\bf 37.0&59.6&40.0&39.3&46.3\\
		\end{tabular}
		\caption{\textbf{Time disentangled \emph{vs.} Time shared - query.} For each tracklet query, the time disentangled scheme uses $T$ embeddings, while the time shared scheme only uses one embedding.}
		\label{tab:ablation6}
	\end{subtable}
	
	\begin{subtable}{0.5\linewidth}
		\centering
		\vspace{-3.5mm}
		\tablestyle{2.5pt}{1.05}
		\captionsetup{width=0.9\linewidth}
		\begin{tabular}{c|c|c|c|cccc}
		\rowcolor{mygray}
		Method&Train Aug.&Epopchs&$\;\;$AP$\;\;$&AP$_{50}$&AP$_{75}$&AR$_{1}$&AR$_{10}$\\
		\shline
		VisTR~\cite{vistr}&random crop&$\sim$500&35.6 & 56.8 & 37.0 & 35.2 & 40.2\\
		EfficientVIS&\xmark&33&\bf 37.0&59.6&40.0&39.3&46.3\\
		EfficientVIS&multi-scale&33&\bf37.9 &59.7 & 43.0 & 40.3 & 46.6
		\end{tabular}
		\caption{\textbf{Convergence speed. (EfficientVIS \emph{vs.} VIS Transformer).} $T=36$ for both VisTR and EfficientVIS. VisTR is equipped with random cropping training augmentation by default.}
		\label{tab:ablation8}
	\end{subtable}
	\begin{subtable}{0.5\linewidth}
	\centering
	\vspace{-0.5mm}
	\captionsetup{width=0.9\linewidth}
	\begin{tabular}{c|c|cccc}
		\rowcolor{mygray}
		Video Frame Rate&AP&AP$_{50}$&AP$_{75}$&AR$_{1}$&AR$_{10}$\\
		\shline
		Original FPS&\bf 35.3&57.6&38.5&37.7&43.5\\
		1.5 FPS&\bf 35.3&57.4&39.1&37.5&42.8\\
	\end{tabular}
	\caption{\textbf{Tracking in low frame rate videos ($T=9$).} We downsample the frame rate of the original YouTube-VIS videos to 1.5 FPS. EfficientVIS is not affected by low video frame rate or dramatic object motions.}
	\label{tab:ablation7}
	\end{subtable}
	\vspace{-3mm}
	\caption{\textbf{Ablation studies} on the YouTube-VIS 2019 \texttt{val} set.}\label{tab:ablation}
	\vspace{-4mm}
\end{table*}

\vspace{-1mm}
\myparagraph{Architecture Settings:} Following~\cite{vistr}, the default video clip length and the number of tracklet queries are set to $T=36$ and $N=10$, respectively. Following~\cite{sun2021sparse,QueryInst,vistr}, the number of iterations is $M=6$ and the default CNN backbone is ResNet-50. We use the above default settings for all experiments unless otherwise specified. Image frame size and pretrained model are the same as~\cite{vistr,Yang_2021_ICCV,QueryInst,cao2020sipmask}.
	
\vspace{-1mm}	
\myparagraph{Training:} EfficientVIS is trained by AdamW with an initial learning rate of $2.5\times10^{-5}$. We train the model for $33$ epochs where the learning rate is dropped by a factor of $10$ at the $27$-th epoch. For example, the model can be trained in 12 hours with 4 RTX 3090 GPUs on YouTube-VIS 2019. The correspondence learning is enabled if the maximum frame number of videos in a dataset is larger than $T$. No training data augmentation is used unless otherwise specified. 

\vspace{-1mm}
\myparagraph{Fully end-to-end Inference:} EfficientVIS does not include any data association or post-processing. \textbf{Reason:} Target tracklet \textbf{within} a clip is generated as a whole in a single forward-pass thanks to the implicit association of our FTSA (Sec.~\ref{subsec:Architecture}). Target tracklets \textbf{between} two clips are automatically correlated owing to our correspondence learning (Sec.~\ref{subsec:Correspondence_learning}). The one-to-one matching loss (Sec.~\ref{subsec:Training}) further enables our method to get rid of post-processing like NMS.
	
\subsection{Ablation Studies}
\vspace{-2mm}
\myparagraph{Time Disentangled Query:} As described in Sec.~\ref{subsec:Architecture}, our tracklet query is designed to be disentangled in time, \ie each query contains $T$ embeddings instead of one embedding. As shown in Table~\ref{tab:ablation6}, we compare it with the time shared query scheme that only uses one single embedding for each tracklet query. Our time disentangled query achieves 1.5 AP higher than the time shared one. We argue that this is because objects in different time may have dramatic appearance changes like heavy occlusion or motion blur. It is more reasonable to use multiple embeddings to separately encode different appearance patterns of an instance.

\vspace{-1mm}
\myparagraph{Factorised Temporo-Spatial Self-Attention:} To evaluate our factorised temporo-spatial self-attention (FTSA), we experiment with different self-attention schemes as shown in Table~\ref{tab:ablation1}. ``S" indicates the spatial self-attention that only performs multi-head self-attention (MSA) within the same frame across the embeddings of different tracklet queries. ``T" denotes the temporal self-attention that only performs MSA within the same tracklet query across the embeddings of different time. ``Joint T-S" is the joint temporo-spatial self-attention that performs MSA across all $T\times N$ embeddings as described in Sec.~\ref{subsec:Architecture}.  We see from the table that the FTSA and ``Joint T-S" achieve better performance than using either ``S" or ``T" standalone. This is because the temporal self-attention can let a query in different time exchange information so as to associate the same instance, while the spatial self-attention models different instances relations in space. Such a query communication in both space and time is necessary for the VIS task.  Compared to ``Joint T-S", our FTSA yields better VIS results. We think the reason is that all attendees in FTSA are strongly related to each other. For example, the embeddings in temporal self-attention are all from the same tracklet query, while those in spatial self-attention are all from the same frame. However, by mixing up all $T\times N$ embeddings in a single self-attention pass, much less relevant information from other tracklets at far temporal positions is directly involved during ``Joint T-S".

\begin{figure}
	\centering
	\includegraphics[width=1\linewidth]{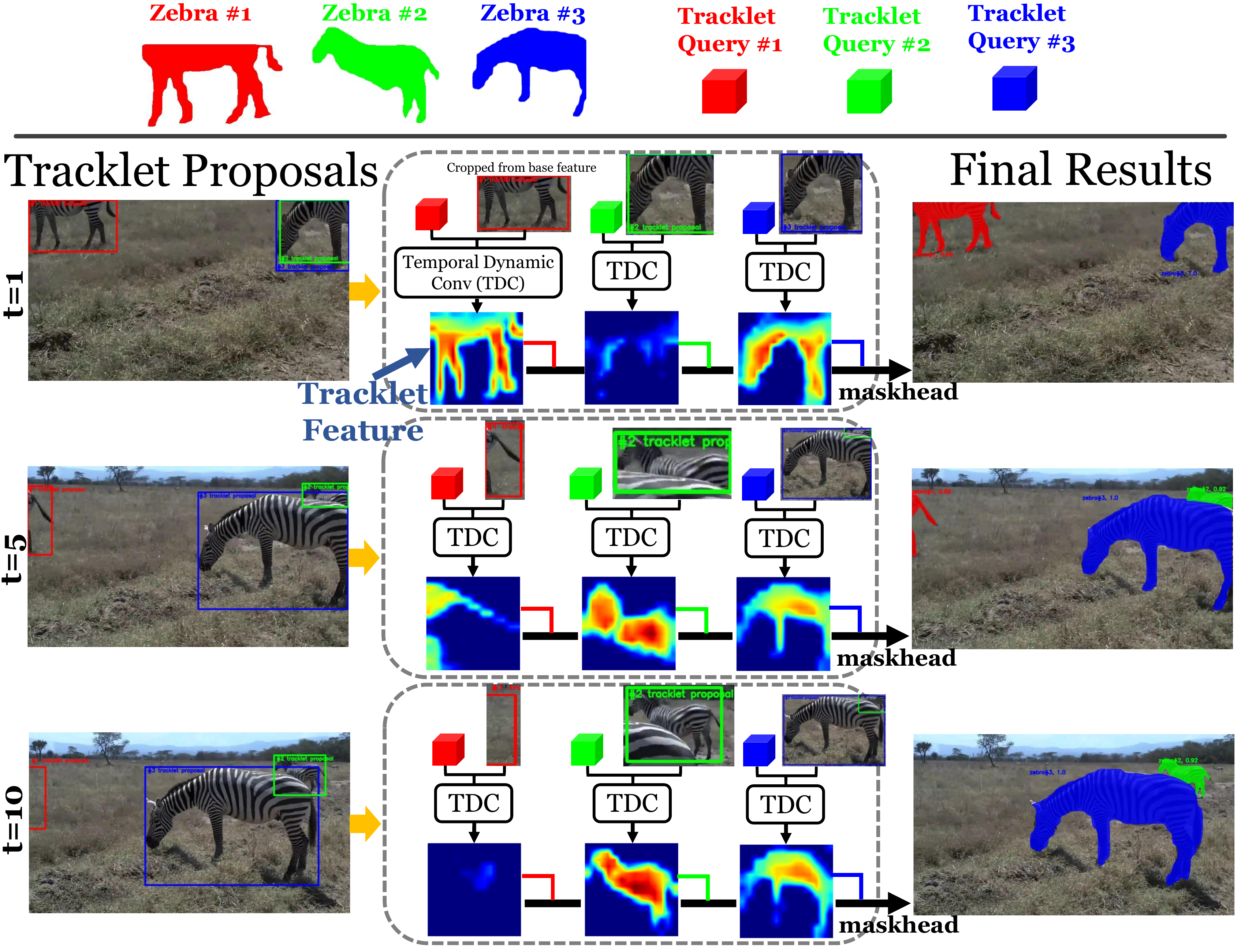}
	\vspace{-8mm}
	\caption{\textbf{Tracklet features carry exclusive target instance information.} The dynamic conv filter conditioned on tracklet query activates tracklet feature only in the region where the target instance appears. The {\#2} tracklet feature at $t$=1 is suppressed by the filter of \textbf{\textcolor{green}{{\#2} zebra}} query even the proposal has drifted to the \textbf{\textcolor{blue}{{\#3} zebra}}.}
	\label{fig:fig4}
	\vspace{-4mm}
\end{figure}

\vspace{-2mm}	
\myparagraph{Temporal Dynamic Convolution:} To assess our temporal dynamic convolution (TDC), we compare it with the still-image dynamic convolution that performs 2D convolution on the current frame only. As shown in Table~\ref{tab:ablation2}, our temporal dynamic convolution improves 1 AP over the still-image version. It suggests that the temporal object context from nearby frames is also an informative cue. Moreover, we visualize the tracklet feature. As shown in Fig.~\ref{fig:fig4}, tracklet feature is highly activated by our TDC in the region where the target instance appears. We also see that tracklet feature is suppressed when the target instance disappears, even if its tracklet proposal has drifted to background clutter or uncorrelated instance. This demonstrates the dynamic filter conditioned on tracklet query is target-specific, which exclusively collects the query's target information from the video base feature yet discards irrelevant information.
	
\vspace{-1mm}	
\myparagraph{Video Clip Length:} As shown in Table~\ref{tab:ablation3}, we experiment with different video clip lengths. EfficientVIS improves 1.7 AP by increasing the number of frames from 9 to 36. It is reasonable because more frames can provide richer temporal context which is important to video tasks. Moreover, since our method runs clip-by-clip in 36 FPS as shown in Table~\ref{tab:ytvis19}, it can be regarded as a near-online fashion with a delay of $\frac{T}{36}$. For example, there is a 0.25s delay when $T=9$.
	
\vspace{-1mm}	
\myparagraph{Correspondence Learning:} To validate the effectiveness of our correspondence learning, we train an EfficientVIS without the correspondence learning. Concretely, we treat every clip as an individual video and do not use tracklet queries from other clips as input during training. In inference, we keep our fully end-to-end paradigm, \ie the output tracklet queries from one clip are used as the initial queries of the next clip. As shown in Table~\ref{tab:ablation5}, the performance significantly drops if EfficientVIS is trained without the correspondence learning. It suggests that our correspondence learning is the key to enabling tracklet query to be a video-level instance representation that can be shared among different video clips.
	
\begin{table*}[h]
	\vspace{-2mm}
	\begin{center}	
		\small
		\begin{tabular}{l|c|c|cc|c|cccc}
			\rowcolor{mygray}
			Method&Publication&Augmentations&Backbone&FPS&AP&AP$_{50}$&AP$_{75}$&AR$_{1}$&AR$_{10}$\\ 
			\shline
			MaskTrack R-CNN~\cite{yang2019video} & ICCV'19 &\xmark & ResNet-50 & 33 & 30.3 & 51.1 & 32.6 & 31.0 & 35.5 \\
			SipMask~\cite{cao2020sipmask} & ECCV'20 &\xmark & ResNet-50 & 34 & 32.5 & 53.0 & 33.3 & 33.5 & 38.9 \\
			CompFeat~\cite{fu2020compfeat} & AAAI'21 & \xmark& ResNet-50 & $<$33 & 35.3 & 56.0 & 38.6 & 33.1 & 40.3 \\
			TraDeS~\cite{wu2021track} & CVPR'21 &\xmark & ResNet-50 & 26 & 32.6 & 52.6 & 32.8& 29.1 & 36.6 \\
			QueryInst~\cite{QueryInst}& ICCV'21 &\xmark & ResNet-50 & 32 &34.6 & 55.8  &36.5  &35.4 & 42.4 \\
			CrossVIS~\cite{Yang_2021_ICCV}& ICCV'21 &\xmark & ResNet-50 & 40 &34.8& 54.6& 37.9 &34.0& 39.0 \\
			VisSTG~\cite{wang2021end2}& ICCV'21 &\xmark & ResNet-50 & 22 &35.2 &55.7& 38.0& 33.6& 38.5 \\
			\textbf{EfficientVIS (Ours)}&CVPR'22 &\xmark& ResNet-50 & 36 & \bf 37.0 & \bf 59.6 & \bf 40.0 & \bf \textcolor{myred2}{39.3} & \bf \textcolor{myred2}{46.3} \\
			
			\hline
			STMask~\cite{li2021spatial} & CVPR'21 &DCN backbone~\cite{dai2017deformable}& ResNet-50 & 29 &33.5 &52.1 &36.9 &31.1 &39.2 \\
			SG-Net~\cite{liu2021sg} & CVPR'21 &multi-scale training& ResNet-50 & 23 &34.8 &56.1 &36.8 &35.8 &40.8 \\
			VisTR~\cite{vistr} & CVPR'21 &random crop training& ResNet-50 & 30 & 35.6 & 56.8 & 37.0 & 35.2 & 40.2 \\
			QueryInst~\cite{QueryInst}& ICCV'21 &multi-scale training & ResNet-50 & 32 &36.2& 56.7& 39.7 &36.1& 42.9 \\
			CrossVIS~\cite{Yang_2021_ICCV}& ICCV'21 &multi-scale training & ResNet-50 &40 &36.3 &56.8 &38.9 &35.6& 40.7\\
			VisSTG~\cite{wang2021end2}& ICCV'21 &multi-scale training & ResNet-50 &22 &36.5 &58.6 &39.0 &35.5& 40.8\\
			\textbf{EfficientVIS (Ours)}&CVPR'22 &multi-scale training& ResNet-50 & 36 & \bf 37.9 & \bf 59.7 & \bf 43.0 & \bf \textcolor{myred2}{40.3} & \bf \textcolor{myred2}{46.6} \\
				
			\hline
			\hline
			MaskTrack R-CNN~\cite{yang2019video} & ICCV'19 &\xmark & ResNet-101 & 29& 31.9& 53.7 &32.3 &32.5& 37.7\\
			SRNet~\cite{ying2021srnet}& ACMMM'21 &\xmark& ResNet-101&35&32.3 &50.2& 34.8& 32.3& 40.1\\
			CrossVIS~\cite{Yang_2021_ICCV}& ICCV'21 &\xmark & ResNet-101 &36& 36.6& 57.3& 39.7& 36.0& 42.0\\
			\textbf{EfficientVIS (Ours)}&CVPR'22 &\xmark& ResNet-101 & 32 & \bf 38.7 & \bf 61.3 & \bf 44.0 & \bf \textcolor{myred2}{40.6} & \bf \textcolor{myred2}{47.7}\\
			
			\hline
			SipMask~\cite{cao2020sipmask} & ECCV'20 &multi-scale training& ResNet-101 & 24 &35.8 &56.0 &39.0 &35.4 &42.4 \\
			STMask~\cite{li2021spatial} & CVPR'21 & DCN backbone~\cite{dai2017deformable} & ResNet-101 & 23 &36.8& 56.8& 38.0& 34.8& 41.8 \\
			SG-Net~\cite{liu2021sg} & CVPR'21 &multi-scale training& ResNet-101 & 20 &36.3& 57.1& 39.6& 35.9 &43.0 \\
			VisTR~\cite{vistr} & CVPR'21 &random crop training& ResNet-101 & 28 & 38.6 & 61.3 & 42.3 & 37.6 & 44.2 \\
			\textbf{EfficientVIS (Ours)}&CVPR'22 &multi-scale training& ResNet-101 & 32 & \bf 39.8 & \bf 61.8 & \bf 44.7 & \bf \textcolor{myred2}{42.1} & \bf \textcolor{myred2}{49.8} \\
			\shline
		\end{tabular}
	\end{center}
	\vspace{-5mm}
	\caption{\textbf{Comparison with real-time state-of-the-art methods on the YouTube-VIS 2019 \texttt{val} set.}}
	\label{tab:ytvis19}
	\vspace{-1mm}
\end{table*}
	
\begin{figure*}
	\centering
	\vspace{-1mm}
	\includegraphics[width=1\linewidth]{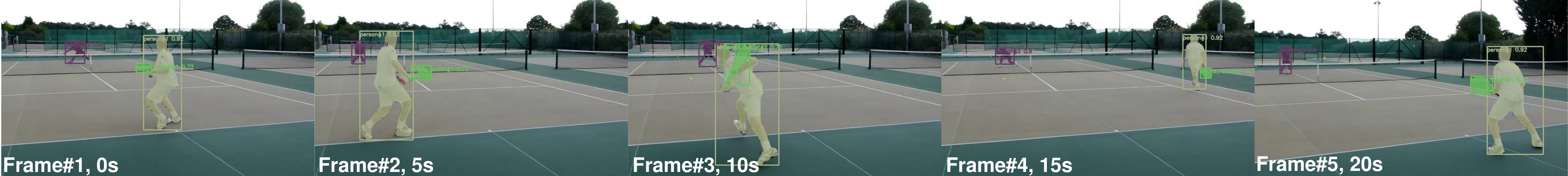}
	\vspace{-6mm}
	\caption{\textbf{Video instance segmentation in a 0.2 FPS video.} EfficientVIS successfully tracks the instances with dramatic movements.}
	\label{fig:fig5}
	\vspace{-3mm}
\end{figure*}
	
\begin{table}[h]
	\vspace{-2mm}
	\begin{center}	
		\footnotesize
		\setlength{\tabcolsep}{3pt}
		\begin{tabular}{l|c|c|cccc}
			\rowcolor{mygray}
			Method&Publication&AP&AP$_{50}$&AP$_{75}$&AR$_{1}$&AR$_{10}$\\ 
			\shline
			MaskTrack R-CNN~\cite{yang2019video} & ICCV'19&28.6& 48.9& 29.6&26.5& 33.8\\
			SipMask$^{\star}$~\cite{cao2020sipmask}&ECCV'20& 31.7 &52.5& 34.0& 30.8& 37.8\\
			CrossVIS~\cite{Yang_2021_ICCV}&ICCV'21&33.3& 53.8 &37.0 &30.1 &37.6\\
			\textbf{EfficientVIS (Ours)} &CVPR'22 &\bf 34.0 & \bf 57.5 & \bf 37.3 & \bf 33.8 & \bf 42.5\\
			\hline
		\end{tabular}
	\end{center}
	\vspace{-5mm}
	\caption{\textbf{Comparison with real-time state-of-the-art methods on the YouTube-VIS 2021 \texttt{val} set.} $\star$ denotes multi-scale training.}
	\label{tab:ytvis21}
	\vspace{-4mm}
\end{table}
	
\vspace{-1mm}
\myparagraph{Fully End-to-end Learnable Framework:} Thanks to the correspondence learning, EfficientVIS fully end-to-end performs VIS by taking tracklet queries from one clip as the initial queries of the next clip. To evaluate this fully end-to-end framework, we compare it with two partially end-to-end schemes, ``hand-craft'' and VisTR~\cite{vistr} in Table~\ref{tab:ablation4}. These two methods perform VIS end-to-end within each clip but require a hand-crafted data association to link tracklets between clips. For the ``hand-craft'' scheme, all the model architectures keep the same as EfficientVIS except that tracklets between two clips are linked by a human-designed rule: 1) We first construct an affinity matrix by taking into account box IoU, box L1 distance, mask IoU, and class label matching like prior tracking works~\cite{wu2021track,zhou2020tracking}; 2) We then solve the tracklet association by the Hungarian algorithm. As shown in Table~\ref{tab:ablation4}, our fully end-to-end scheme performs better than the partially end-to-end framework in different $T$ settings while greatly simplifying the previous VIS paradigms. The reason for this performance gap is that the tracklet linking between clips in hand-crafted scheme tends to be sub-optimal, while that in our fully end-to-end framework is learned and optimized by ground truths during training.

\vspace{-1mm}
\myparagraph{Tracking in Low Frame Rate Videos:} 
Tracking object instances in low frame rate videos is a hard problem, where motion-based trackers usually fail. Motion-based trackers suppose instances move smoothly over time and impose spatial movement constraints to prevent trackers from associating very distant instances. In contrast, our tracking is determined by instance appearance rather than motion, as we retrieve a target instance solely conditioned on its query representation regardless of the target spatial distance among different frames. Therefore, our tracking is not limited by dramatic instance movements. To demonstrate this property, we downsample the frame rate of the YouTube-VIS dataset to 1.5 FPS and test our method. As shown in Table~\ref{tab:ablation7}, EfficientVIS successfully maintains its VIS performance in the low frame rate video setting. As shown in Fig.~\ref{fig:fig5}, we also visualize the results of EfficientVIS on a very low frame rate video whose frame is sampled every 5 seconds, \ie 0.2 FPS.

\vspace{-1mm}
\myparagraph{Fast Convergence:} In our method, we crop video using tracklet proposal and collect target information from the proposal. Compared to VIS transformer (VisTR)~\cite{vistr} that uses frame-wise dense attention, this RoI-wise design eliminates much background clutter and redundancies in videos and enforces EfficientVIS to focus more on informative regions, making our convergence 15$\times$ faster as shown in Table~\ref{tab:ablation8}. This RoI-wise pipeline also leads to better accuracy than VisTR. This is because tracklet proposal can avoid regions outside the target being segmented (second figure in appendix), which results in more precise masks and significantly higher AP$_{75}$ as evidenced in Table~\ref{tab:ablation8}.

\vspace{-1mm}
\subsection{Comparison to State of the Art}
\label{subsec:sota}
\vspace{-3mm}
\myparagraph{YouTube-VIS 2019:} In Table~\ref{tab:ytvis19}, we compare EfficientVIS with the real-time state-of-the-art VIS models. EfficientVIS is the only fully end-to-end framework while achieving superior accuracy. We attribute the strong performance to two main aspects: 1) Our object tracking is learned via the end-to-end framework; 2) Our clip-by-clip processing and temporal dynamic convolution enrich temporal object context that is helpful for handling occlusion or motion blur. As shown in Table~\ref{tab:ytvis19}, we \emph{achieve notably higher recall} (AR) than other competitors. We think the reason for the high recall is that we recognize more heavily occluded or blurred objects, which are usually missed by common methods. 

For a fair comparison, Table~\ref{tab:ytvis19} only lists real-time methods using single model without extra data. There are two works with 40+ AP, MaskProp~\cite{bertasius2020classifying} ($<$5.6 FPS) and Propose-Reduce~\cite{lin2021video} (1.8 FPS), which however are not real-time and adopt multiple models or extra training data. MaskProp uses DCN backbone~\cite{dai2017deformable}, HTC detector~\cite{chen2019hybrid}, extra High-Resolution Mask Refinement network, etc. Propose-Reduce is trained with additional DAVIS-UVOS~\cite{caelles20192019} and COCO pseudo videos datasets. These elaborated implementations are beyond the scope of our work, as our main goal is to present a simple end-to-end model with real-time inference and fast training. 

\vspace{-1mm}
\myparagraph{YouTube-VIS 2021:} We experiment with EfficientVIS on YouTube-VIS 2021 in Table~\ref{tab:ytvis21}. EfficientVIS achieves state-of-the-art AP without bells and whistles. Similar to the findings on YouTube-VIS 2019, EfficientVIS significantly improves AR over other state of the arts.

\vspace{-2mm}	
\section{Conclusion}
\vspace{-2mm}
This work presents a new VIS model, EfficientVIS, which simultaneously classifies, segments, and tracks multiple object instances in a single end-to-end pass and clip-by-clip fashion. EfficientVIS adopts tracklet query and proposal to respectively represent instance appearance and position in videos. An efficient query-video interaction is proposed for associating and segmenting instances in each clip. A correspondence learning is designed to correlate instance tracklets between clips without data association. The above designs enable a fully end-to-end framework achieving state-of-the-art VIS accuracy with fast training and inference.

\myparagraph{Acknowledgement.} This paper is supported in part by a gift grant from Amazon Go and National Science Foundation Grant CNS1951952.

{\small
	\bibliographystyle{ieee_fullname}
	\bibliography{EfficientVIS}
}

\clearpage
\newpage
\appendix

\begin{strip}
\centering
  \includegraphics[width=\textwidth]{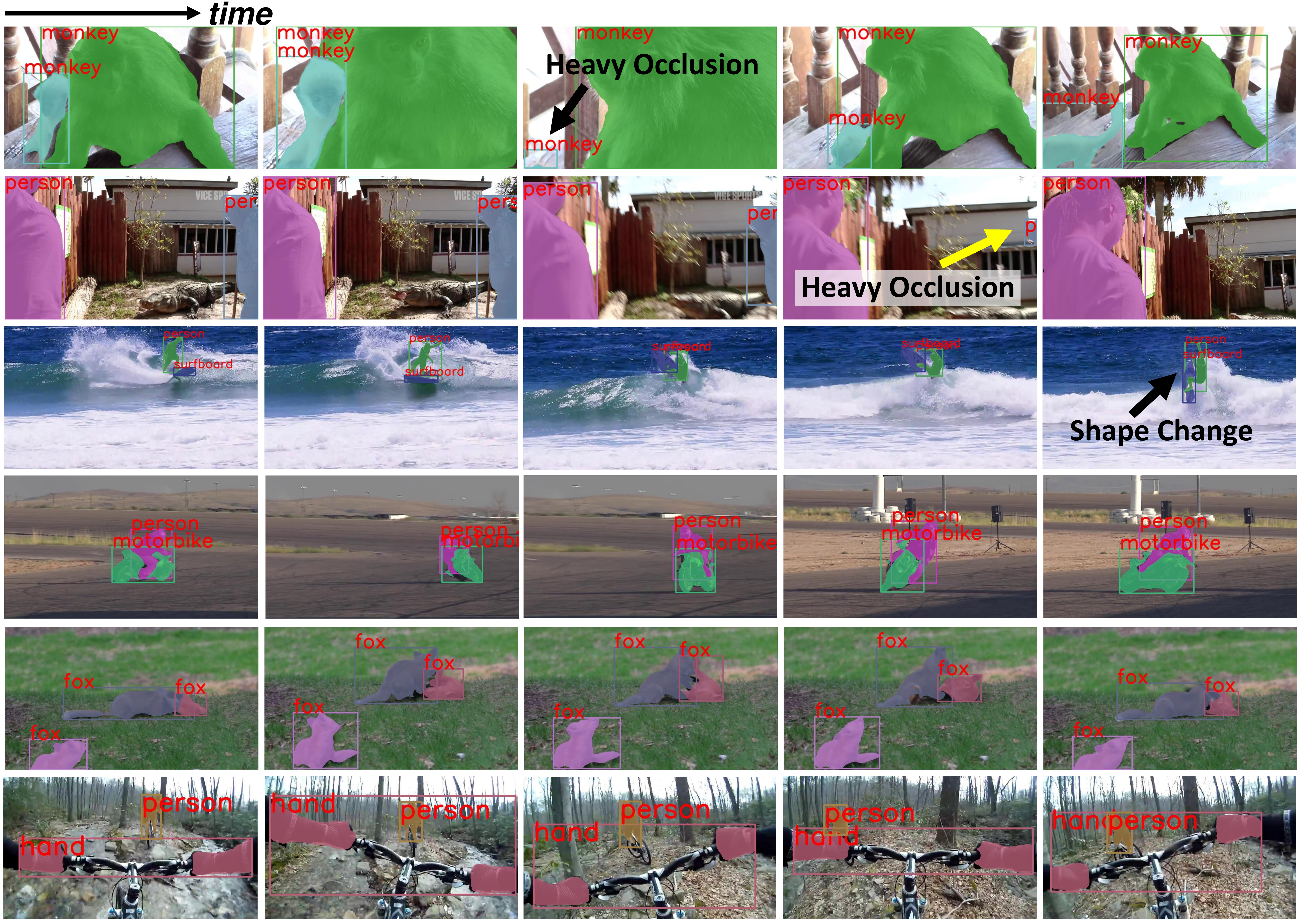}
  \captionof{figure}{\textbf{Visualization on the YouTube-VIS dataset.} One color indicates one identity. The boxes denote our tracklet proposals.}
  \label{fig:fig6}
\end{strip}

\begin{center}{\bf \Large Appendix}\end{center}\vspace{-2mm}

\section{Qualitative Results on YouTube-VIS}
As shown in Fig.~\ref{fig:fig6}, we visualize our video instance segmentation results on the YouTube-VIS 2019 dataset. We see from the figure that EfficientVIS can well recognize heavily occluded object instances. The reason is that the target instance information is propagated back and forth in a clip thanks to our clip-by-clip processing and \emph{temporal dynamic convolution}. In this way, the non-occluded instance appearances from nearby frames are propagated to provide strong cues to recognize those heavily occluded instances. We also see in the figure that our tracklet proposal can successfully track target instance even its shape dramatically changes over time. This is because tracklet proposal is regressed conditioned on the target query representation, and we do not impose space-time constraints or smoothness. Thus, tracklet proposal is not limited by the target positions or shapes in nearby frames.

\section{Qualitative Comparison}
\begin{figure*}
	\centering
	\includegraphics[width=1\linewidth]{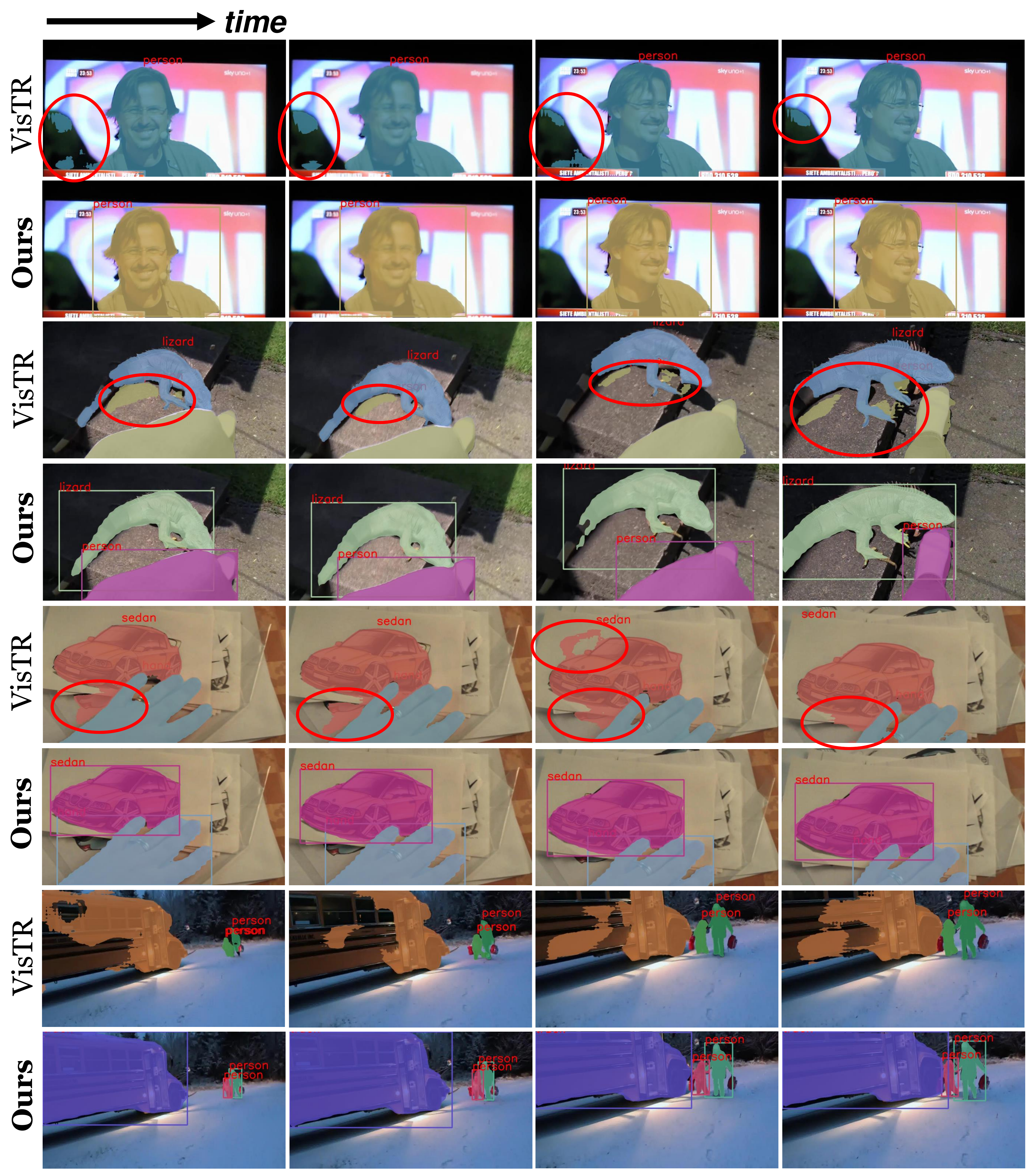}
	\caption{\textbf{Qualitative comparison} with the VIS transformer (VisTR)~\cite{vistr}.}
	\label{fig:fig7}
	\vspace{-2mm}
\end{figure*}

As shown in Fig.~\ref{fig:fig7}, we compare EfficientVIS with the VIS transformer (VisTR)~\cite{vistr}. Since VisTR produces an instance mask by segmenting the whole frame, one instance mask may easily contaminate other instances or background regions as shown in the figure. This suggests that it is hard to enforce the query representation in VisTR to be very discriminative to distinguish target object instance from the whole scene. In contrast to this frame-wise scheme, EfficientVIS filters out many irrelevant instances and regions by our tracklet proposals. Our method only needs to enforce tracklet query to distinguish target instance from the proposal region, which is easier for the model to achieve.

\end{document}